\documentclass[journal]{IEEEtran}
\usepackage{ifpdf}
\usepackage[colorlinks,linkcolor=red,anchorcolor=blue,citecolor=green]{hyperref}
\usepackage[pdftex]{graphicx}
\usepackage{amsmath}
\usepackage{algorithmic}
\usepackage{array}
\usepackage[caption=false,font=normalsize,labelfont=sf,textfont=sf]{subfig}
\usepackage{fixltx2e}
\usepackage{dblfloatfix}
\usepackage{url}

\usepackage{times}
\usepackage{epsfig}
\usepackage{amssymb}
\usepackage{bm}
\usepackage{algorithm}
\usepackage{verbatim}
\usepackage{color}
\usepackage{authblk}

\begin{document}
	
\title{Landmark Detection and 3D Face Reconstruction for Caricature using a Nonlinear Parametric Model}

\author{Hongrui~Cai,\quad
	 Yudong~Guo,\quad
	 Zhuang~Peng,\quad
	 Juyong~Zhang*\thanks{*Email: {\texttt{juyong@ustc.edu.cn}}.}
\IEEEcompsocitemizethanks{\IEEEcompsocthanksitem J. Zhang, Y. Guo and Z. Peng are with School of Mathematical Sciences, University of Science and Technology of China.
\IEEEcompsocthanksitem H. Cai is with School of Data Science, University of Science and Technology of China.
}
}


\maketitle

\begin{abstract}
Caricature is an artistic abstraction of the human face by distorting or exaggerating certain facial features, while still retains a likeness with the given face. Due to the large diversity of geometric and texture variations, automatic landmark detection and 3D face reconstruction for caricature is a challenging problem and has rarely been studied before. In this paper, we propose the first automatic method for this task by a novel 3D approach. To this end, we first build a dataset with various styles of 2D caricatures and their corresponding 3D shapes, and then build a parametric model on vertex based deformation space for 3D caricature face. Based on the constructed dataset and the nonlinear parametric model, we propose a neural network based method to regress the 3D face shape and orientation from the input 2D caricature image. Ablation studies and comparison with state-of-the-art methods demonstrate the effectiveness of our algorithm design. Extensive experimental results demonstrate that our method works well for various caricatures. Our constructed dataset, source code and trained model are available at \url{https://github.com/Juyong/CaricatureFace}.
\end{abstract}

\IEEEpeerreviewmaketitle

\section{Introduction}

\begin{figure}
	\begin{center}
		\includegraphics[width=1.0\linewidth]{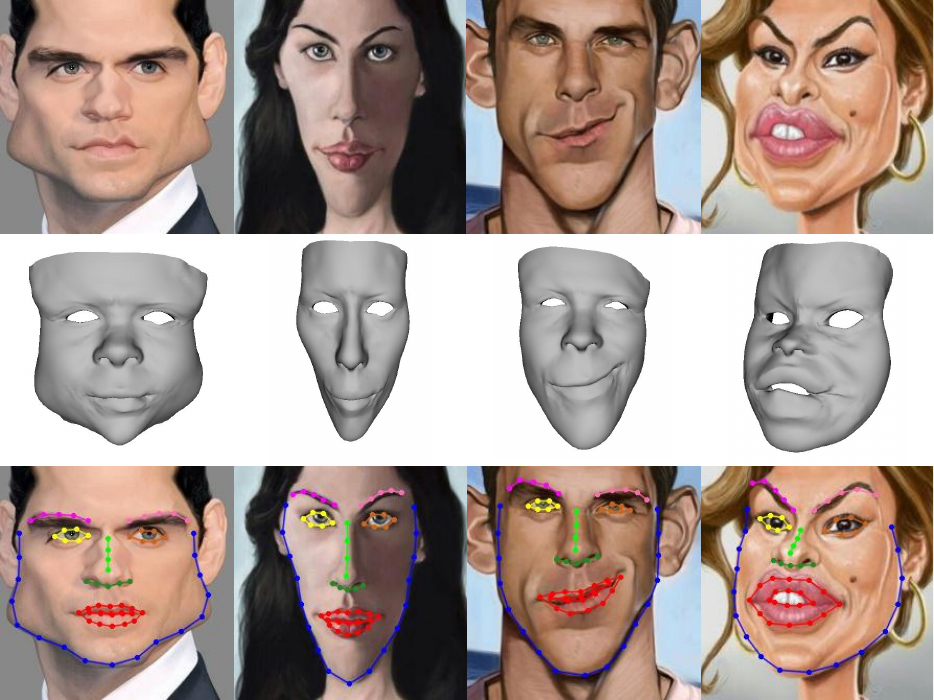}
	\end{center}
	\caption{Some examples of automatic landmark detection and 3D reconstruction on test set. Given a single caricature image (first row), our algorithm generates its 3D model with orientation (second row) and corresponding 68 landmarks (third row).}
	\label{fig:showpic}
\end{figure}

As a vivid artistic form that represents human faces in abstract and exaggerated ways, caricature is mainly used to express satire and humor for political or social incidents. It also has many applications in our daily life, such as social network, animation and entertainment industry. Since Brennan developed the first caricature generator in 1985~\cite{brennan1985caricature}, the studies of caricatures have mainly focused on some specific tasks, such as caricature generation~\cite{liang2002example,shumsample,cao2018carigans,shi2019warpgan}, caricature recognition~\cite{klare2012towards,ouyang2014cross,abaci2015matching}, and  caricature reconstruction~\cite{lewiner2011interactive,vieira2013three,han2017deepsketch2face,wu2018alive}. Most of these tasks need facial landmarks to help to preprocess the caricatures. As a fundamental process for various caricature processing tasks, automatic facial landmark detection and 3D face reconstruction can greatly improve the efficiency and accuracy of other caricature processing tasks. Although the state-of-the-art face alignment methods work well for normal facial images, they are not applicable to caricatures. For example, it still needs to manually refine the landmark positions after applying face alignment methods to caricature images, as reported in~\cite{huo2017webcaricature,wu2018alive,cao2018carigans}.

Compared with other tasks like caricature generation~\cite{cao2018carigans,shi2019warpgan,chu2020learning} and editing~\cite{chen2020modeling}, there is little research on automatic landmark detection for caricatures. As far as we know, one related work is proposed by Sadimon and Haron~\cite{sadimon2015neural}, which adopted the neural network to predict a facial caricature configuration. However, it can not process a single 2D caricature without its original facial image because the training dataset is constructed by image pairs- one normal facial image and its corresponding caricature image. Besides, their training and testing caricatures are all from exactly one artist, and thus the trained model can not be adapted to other caricatures with different art styles. There exist two main difficulties of facial landmark detection for caricature. One difficulty is that caricatures have abstract and exaggerate patterns, and another is that caricatures have large representation varieties among different artists. As pointed out in~\cite{huo2017webcaricature}, compared with landmark detection on normal facial images, it is much more challenging on landmark detection for caricatures.

In comparison to normal facial images, caricatures have two fundamental attributes- exaggeration and variety, and thus approaches for standard landmark detection can not be directly applied to solve this problem. One straightforward way is to regress the 2D landmarks' coordinates of caricature directly. However, 2D landmarks are controlled by facial shape, expression, orientation, and artistic style, which makes it a challenging problem to detect 2D landmarks. In order to alleviate the problem difficulty, we propose to decouple these factors. By regressing the 3D face model and orientation, 2D landmarks can be recovered by projecting the 3D landmarks with the orientation.

However, existing parametric 3D face models are mainly designed to represent normal face shapes, and thus they do not work well for caricature faces due to their limited capability of extrapolation. In this paper, to solve this challenging problem, we specifically design a parametric model for 3D caricature faces and propose a method for landmark detection and 3D reconstruction of caricature based on this model. To this end, we manually label landmarks of about 6K caricature images with different styles. We further automatically generate nearly 2K caricatures with labeled landmarks from normal facial images via the method described in~\cite{cao2018carigans}. Based on the labeled landmarks, we recover the corresponding 3D caricature shape and orientation using an optimization method. With the large scale training dataset, we propose a novel convolutional neural network based method to regress the 3D caricature shape and orientation from the input 2D caricature. To well represent the 3D exaggerated face, we propose to regress its deformation representation rather than the Euclidean coordinates, which helps to improve the landmark detection and 3D reconstruction ability. In summary, the main contributions of this paper include the following aspects:
\begin{itemize}
	\item To the best of our knowledge, this is the first work for automatic landmark detection and 3D face reconstruction for general caricatures.
	\item Rather than directly regress the 2D landmarks, we regress the 3D caricature shape and orientation from input 2D caricature image. 3D caricature shape is represented by a nonlinear parametric model learned from our constructed 3D caricature dataset.
\end{itemize}

Comparisons with state-of-the-art methods and ablation studies demonstrate the effectiveness of our algorithm pipeline and each module of our proposed method. Extensive qualitative and quantitative experiments demonstrate that our method can automatically produce high accuracy results of 2D landmark detection and 3D shape reconstruction for caricature.
\section{Related Work}

\begin{figure*}
	\begin{center}
		\includegraphics[width=1.0\linewidth]{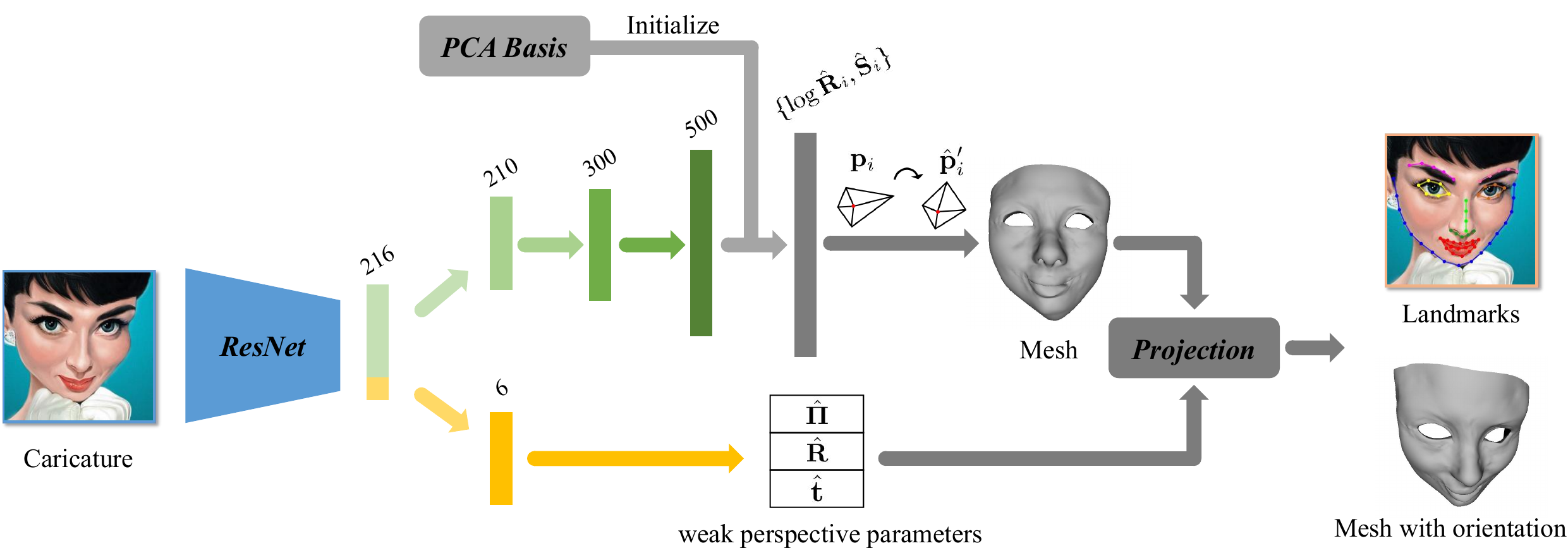}
	\end{center}
	\caption{Overview of our proposed Framework for Landmark Detection and 3D Reconstruction on general caricatures. Our network includes two parts, an encoder and a decoder. We use ResNet-34~\cite{he2016deep} backbone as the encoder and 3 Fully Connected (FC) layers as the decoder to recover the 3D caricature shape. The PCA basis of vertex based deformation presentation ${\{\log \mathbf{R}_i, \mathbf{S}_i\}}$ is used to initialize the last FC layer.}
	\label{fig:pipeline}
\end{figure*}

This section briefly reviews some works related to this paper, with a special focus on face alignment and 3D face reconstruction for normal facial images, and face alignment and 3D face reconstruction for caricatures.

\noindent{\bf Face Alignment.}
Face alignment and landmark detection for normal facial images have achieved great success in the last few years with the power of convolution neural networks. Kazemi and Sullivan~\cite{kazemi2014one} used an Ensemble of Regression Trees to estimate the facial landmark positions, and their method has been integrated into the Dlib library~\cite{king2009dlib}, a modern C++ toolkit containing some machine learning algorithms. Wu \emph{et al.}~\cite{wu2017facial} proposed \emph{vanilla} CNN, which is naturally hierarchical and requires no auxiliary labels beyond landmarks. Kowalski \emph{et al.}~\cite{kowalski2017deep} developed Deep Alignment Network (DAN), a robust deep neural network architecture that consists of multiple stages. By adopting a coarse-to-fine Ensemble of Regression Trees, Valle \emph{et al.}~\cite{valle2018deeply} proposed a real-time facial landmark regression algorithm. Liu \emph{et al.}~\cite{liu2019semantic} noticed that the semantic ambiguity degrades the detection performance and addressed this issue by latent variable optimization methods. Dong \emph{et al.}~\cite{dong2018supervision} presented an unsupervised approach to improving facial landmark detectors, and Honari \emph{et al.}~\cite{honari2018improving} showed a new architecture and training procedure for semi-supervised landmark localization. To solve the occlusion problem, Zhu \emph{et al.}~\cite{zhu2019robust} developed an occlusion-adaptive deep network, which contains a geometry-aware module, a distillation module, and a low-rank learning module. Merget \emph{et al.}~\cite{merget2018robust} proposed a novel network architecture that has an implicit kernel convolution between a local-context subnet and a global-context subnet composed of dilated convolutions.

\noindent{\bf 3D Face Reconstruction from A Single Image.}
3D face reconstruction algorithms can be divided into different categories on the ground of the input modality: single RGB image based~\cite{guo2018cnn}, video based~\cite{liu2017robust}, depth image based~\cite{wang2017joint}, and so on. 3D face reconstruction from a single image is to recover 3D facial geometry from a given facial image, which has applications like face recognition~\cite{blanz2003face,tuan2017regressing}, face alignment~\cite{zhu2016face, feng2018joint} and expression transfer~\cite{thies2015realtime,thies2016face2face}.  Since Blanz and Vetter proposed a 3D Morphable Model (3DMM) in 1999~\cite{blanz1999morphable}, model-based methods have become popular in solving problems of 3D face reconstruction. Earlier, a large number of model-based algorithms considered some significant facial parts between 2D images and 3D templates, such as facial landmarks~\cite{cao2014displaced,jeni2015dense,grewe2016fully,thies2016face2face,JiangZDLL18}, latent representation~\cite{huber2015fitting} and so on. Cao \emph{et al.}~\cite{cao2013facewarehouse} utilized some RGBD sensors to create an extensive face database named FaceWareHouse, which contains 150 identities and 47 expressions of each identity. In recent years, deep learning based methods have shown promising results in terms of computation time, robustness to occlusions, and reconstruction accuracy~\cite{jin20203d}. Guo \emph{et al.}~\cite{guo2018cnn} proposed a real-time dense face reconstruction method by constructing a large scale dataset augmented based on traditional optimization methods and adopting a coarse-to-fine CNN framework. During the same year, Tran \emph{et al.}~\cite{tran2018nonlinear} demonstrated a nonlinear 3DMM, which is learned from a large set of unconstrained face images without collecting 3D face scans. Gecer \emph{et al.}~\cite{gecer2019ganfit} harnessed Generative Adversarial Networks (GANs) for reconstructing facial texture and shape from single images by training a generator of facial texture in UV space. Feng \emph{et al.}~\cite{feng20183d} presented a model-free method to rebuild the 3D facial geometry from a single light field image with a densely connected network. However, due to the diversity of style and geometry of caricatures, the approaches for normal face reconstruction can not be directly applied to general caricatures.

\noindent{\bf Face Alignment and Reconstruction of Caricature.}
Compared with researches on normal facial images, there are fewer works about caricatures~\cite{o1997three,o19993d}. For face reconstruction, existing methods mainly focus on constructing a 3D caricature model from a normal 3D face model. Lewiner \emph{et al.}~\cite{lewiner2011interactive} introduced a caricature tool that interactively emphasizes the differences between two 3D meshes by utilizing the manifold harmonic basis of a shape to control the deformation and scales intrinsically. Vieira \emph{et al.}~\cite{vieira2013three} proposed a method based on deformations by manipulation of moving spherical influence zones. Sela \emph{et al.}~\cite{sela2015computational} presented a framework to scale the gradient fields of the surface coordinates by a function of the Gaussian curvature of the surface and solve a corresponding Poisson equation to find the exaggerated shape. Besides, there are some works on modeling 3D caricatures from images. Liu \emph{et al.}~\cite{liu2009semi} chose a semi-supervised manifold regularization(MR) method to learn a regressive model between 2D normal faces and enlarged 3D caricatures. With the power of deep learning, Han \emph{et al.}~\cite{han2017deepsketch2face} developed a CNN based sketching system that allows users to draw freehand imprecise yet expressive 2D lines representing the contours of facial features. With an intrinsic deformation representation that enables considerable face exaggeration, Wu \emph{et al.}~\cite{wu2018alive} introduced an optimization framework to address this issue. However,~\cite{wu2018alive} needs labeled landmarks as input, which are not easy to get and always need manually labeling. Different from~\cite{wu2018alive}, we propose a learning based approach to automatically detect landmarks and regress 3D shape from the input 2D caricature. Furthermore, our method constructs a nonlinear parametric model based on deformation representation, which greatly improves the reconstruction accuracy. As~\cite{wu2018alive} is the state-of-the-art caricature reconstruction method, we adopt it to construct the 3D caricature shape set. Landmark detection on caricature images is also a fundamental problem of caricature perception, but there exist few works on this topic. As a related research direction, manga images have aroused Stricker \emph{et al.}'s~\cite{stricker2018facial} interest. Based on DAN~\cite{kowalski2017deep} framework, they proposed a new landmark annotation model for manga images and a deep learning approach to detect them.  Huo \emph{et al.}~\cite{huo2017webcaricature} shows that caricature landmark detection is of great interest, but researches on this topic are still far from saturated. Besides, most studies on caricature generation need facial landmarks as control points ~\cite{liang2002example,cao2018carigans,shi2019warpgan}, which demonstrate that facial landmarks play an essential role in caricature related researches.
\section{Algorithm}

Given a 2D caricature, we aim to automatically reconstruct its 3D face shape and obtain landmarks around its eyes, nose, mouse, and so on, as shown in Fig.~\ref{fig:showpic}. To this end, we construct a 2D caricature dataset with around 8K 2D images and their corresponding labeled landmarks. The dataset contains both artists-designed caricatures and machine-generated caricatures. With their corresponding 68 landmarks, we build a 3D caricature dataset via an optimization based method~\cite{wu2018alive}. Then, based on a deformation representation, we propose an encoder-decoder framework to directly recover the 3D face shape and weak perspective parameters from the input 2D caricature image. Notably, we use the principal component analysis (PCA) basis to initialize the weight of the last fully connected layer. The algorithm pipeline is shown in Fig.~\ref{fig:pipeline}. In the following, we give the algorithm details for each component.

\subsection{Dataset Construction and Augmentation}
\label{sec:data}
Currently, there exist some public available caricature datasets. For the study of caricature recognition, Huo \emph{et al.}~\cite{huo2017webcaricature} constructed a WebCaricature database including 6042 caricatures and 5974 photographs from 252 persons with 17 labeled facial landmarks for each image. Mishra \emph{et al.}~\cite{mishra2016iiit} built IIIT-CFW database for face classification and caricature generation, which contains 8928 cartoon faces of 100 public figures with annotation of various attributes, e.g., face bounding box, age group, facial expression, and so on. However, these datasets can not be directly used for our task as they do not supply enough labeled landmarks for 3D reconstruction.

\begin{figure}[!b]
	\begin{center}
		\includegraphics[width=1.0\linewidth]{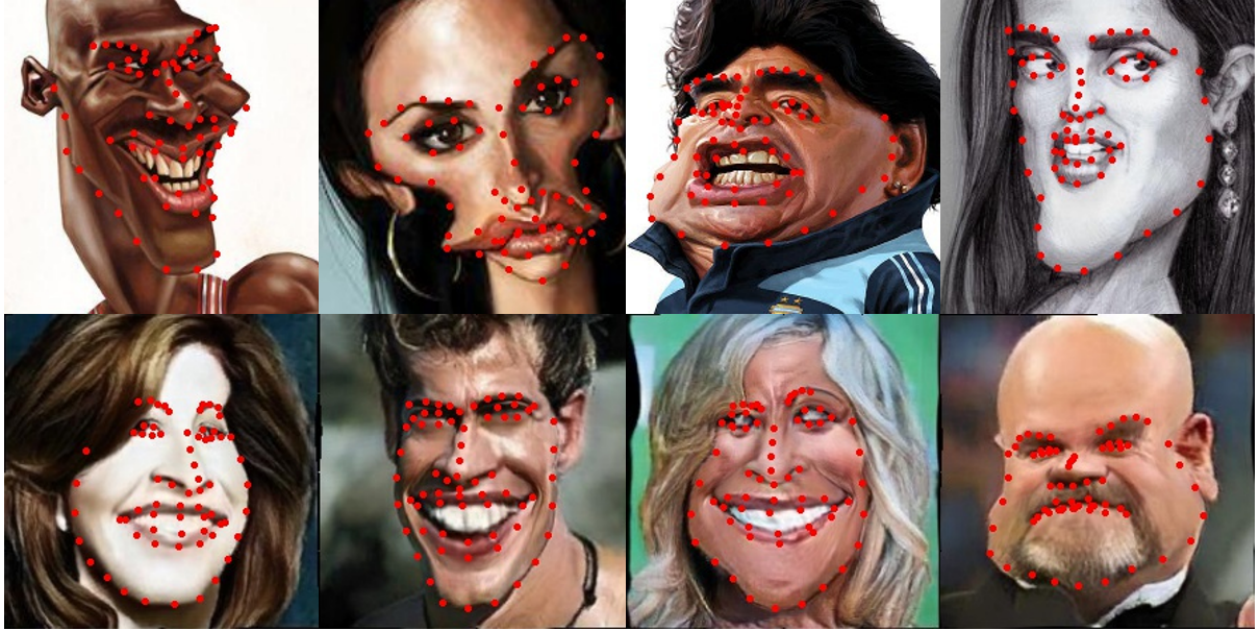}
	\end{center}
	\caption{The first row shows some examples of our collected images with manually labeled landmarks, while the second row shows some examples of our augmented images and corresponding landmarks generated by~\cite{cao2018carigans}.}
	\label{fig:selected}
\end{figure}

By searching and selecting nearly 6K various caricatures from different artists on the Internet, we construct a caricature dataset in which each caricature has 68 labeled landmarks. The landmark positions are initialized via the Dlib library~\cite{king2009dlib}, and then manually refined. To further increase the diversity of our dataset, we design a data augmentation method based on CariGANs~\cite{cao2018carigans}. CariGANs is able to translate normal facial images to caricatures with two generative adversarial networks (GANs), namely CariGeoGAN and CariStyGAN. CariGeoGAN learns a mapping to exaggerate the shape by adjusting facial landmarks, while CariStyGAN learns another mapping to translate the appearance from normal facial image style to caricature style. With trained CariGANs, we can generate a caricature and its corresponding 68 landmarks from a given normal facial image. In this way, we generate around 2K caricatures and add them to our dataset. Some examples of our collected data and augmented data are shown in Fig.~\ref{fig:selected}.

For each 2D caricature, based on the labeled 68 facial landmarks, we adopt an optimization based method~\cite{wu2018alive} to recover its 3D shape. In this way, we build a caricature dataset containing around 8K 2D labeled images and their corresponding 3D shapes. Notably, this dataset contains different geometry shapes and image styles since the 2D caricatures are drawn by different artists or generated by the data augmentation method.

\subsection{Deformation Representation of 3D Caricatures}
\label{sec:construction}
Some statistical parametric models, like 3DMM~\cite{blanz1999morphable} and FaceWareHouse~\cite{cao2013facewarehouse}, are popularly used in 3D face reconstruction to represent a complex face shape with a low dimensional parametric vector. This kind of representation makes optimization and learning based 3D face reconstruction easier. However, linear parametric models are only good for interpolation in the shape space of 3D normal faces but do not work well for extrapolation in 3D caricature shape space. Therefore, to recover exaggerated 3D shapes of various caricatures, we adopt a vertex based deformation representation. Compared with 3D Euclidean coordinates, this deformation representation is suitable to represent local and large deformation in a natural way, which makes the reconstructed exaggerated meshes more natural and match the input 2D caricature quite well.

To make our paper self-contained, we first introduce the deformation representation between two meshes with the same topology.

Suppose there are two meshes with the same topology, which means the number of vertices, the order of vertices, and the connectivity of vertices of them are all identical. We treat one mesh as a reference mesh and another as a target deformed mesh. We denote the position of the $i^{\rm th}$ vertex $v_i$ on the reference as $\mathbf{p}_i$ and the $i^{\rm th}$ vertex $v_i$ on the target as $\mathbf{p}_i'$. We can define the deformation matrix in the one-ring neighborhood of $v_i$ from the reference to the target as an affine transformation matrix $\mathbf{T}_i$ by minimizing
\begin{equation}
\mathop{\min_{\mathbf{T}_{i}}}\sum_{j\in{\mathcal{N}_i}} c_{ij}\|{(\mathbf{p}_{i}^{\prime}-\mathbf{p}_{j}^{\prime})-\mathbf{T}_{i}(\mathbf{p}_{i}-\mathbf{p}_{j})}\|_{2}^2,
\label{eq:vertex_representation}
\end{equation}
where $\mathcal{N}_i$ is the neighborhood index set of $v_i$, and $c_{ij}$ is the cotangent weight~\cite{botsch2007linear} to avoid discretization bias in deformation. With polar decomposition, the deformation matrix $\mathbf{T}_i$ can be decomposed into a rigid component represented by a rotation matrix $\mathbf{R}_i$ and a non-rigid component represented by a real symmetry matrix $\mathbf{S}_i$, as $\mathbf{T}_i = \mathbf{R}_i\mathbf{S}_i$.

To obtain efficient linear combination, we use the axis-angle representation~\cite{diebel2006representing} to replace the rotation matrix $\mathbf{R}_i$. Following Rodrigues' rotation formula, for the $i^{\rm th}$ vertex $v_i$, we denote the cross-product matrix and rotation angle by $\mathbf{K}_i$, $\theta_i$. We can convert $\mathbf{R}_i$ to a matrix logarithm notation:
\begin{equation}
\log \mathbf{R}_i = \theta_i\mathbf{K}_i,
\end{equation}
\begin{equation}
\mathbf{K}_i
={
\left[ \begin{array}{ccc}
0 & -k_{i,z} & k_{i,y} \\
k_{i,z} & 0 & -k_{i,x} \\
-k_{i,y} & k_{i,x} & 0 \\
\end{array} 
\right ]},
\end{equation}
where $\mathbf{k}_i = (k_{i,x},k_{i,y},k_{i,z})\in{\mathbb{R}^3}$ and $\|\mathbf{k}_i\|_2=1$. Then, the logarithm rotation matrix $\log \mathbf{R}_i$ can be represented by a vector $\mathbf{r}_i = \theta_i\mathbf{k}_i\in {\mathbb{R}^3}$ and the scalar matrix $\mathbf{S}_i$ can be represented by a vector $\mathbf{s}_i\in {\mathbb{R}^6}$. To handle the ambiguity of axis-angle representation, Gao \emph{et al.}~\cite{gao2019sparse} propose an integer programming approach to make all $\mathbf{r}_i$ as consistent as possible globally. In this paper, we define $[\mathbf{r}_i, \mathbf{s}_i]\in {\mathbb{R}^{9}}$ as the deformation representation/gradient of the $i^{\rm th}$ vertex $v_i$ on a target mesh, correspondingly, $\{\log \mathbf{R}_i, \mathbf{S}_i\}$ is its matrix form.

The deformation representation has many advantages, especially for our method, it can be used for linear combination~\cite{alexa2002linear} of two rotation matrices $\mathbf{R}_i^{\rm 0}$ and $\mathbf{R}_i^{\rm 1}$ by $\exp(\log \mathbf{R}_i^{\rm 0}+\log \mathbf{R}_i^{\rm 1})$. To build a deformation space, we choose a reference mesh and $n$ deformed meshes which have the same topology with the reference model. For the $i^{\rm th}$ vertex of the $l^{\rm th}$ deformed mesh, we obtain its deformation representation ${\{\log \mathbf{R}_i^l, \mathbf{S}_i^l\}}(l=1,2,\ldots, n)$. Then, corresponding to an essential deformation representation, a target mesh can be approximately reconstructed by a linear combination of several known deformation gradients. In detail, based on a reference mesh and $n$ deformed meshes, we build a linear combination of deformation gradients for the $i^{\rm th}$ vertex $v_i$ as
\begin{equation}
\mathbf{T}_i(\mathbf{w}) = \exp(\sum_{l=1}^n{w_{R,l}}\log \mathbf{R}_i^l){(\mathbf{I} + \sum_{l=1}^nw_{S,l} (\mathbf{S}_i^l-\mathbf{I}))},
\label{eq:DRCombination}
\end{equation}
where $\mathbf{w} = (\mathbf{w}_R, \mathbf{w}_S)$ is the combination weight vector, consisting of weights of rotation $\mathbf{w}_R = \{w_{R,l}|l=1,\cdots,n\}$ and weights of scaling/shear $\mathbf{w}_S = \{w_{S,l}|l=1,\cdots,n\}$.

Given a target mesh, we can calculate its optimal weight $\mathbf{w}$ by minimizing the following energy:
\begin{equation}
\min_{\mathbf{w}}\sum_{v_i\in{\mathcal{V}}}\sum_{j\in{\mathcal{N}_i}} c_{ij}\|{({\mathbf{p}_i^{\prime}}-{\mathbf{p}_j^{\prime}})-\mathbf{T}_{i}(\mathbf{w})(\mathbf{p}_i-\mathbf{p}_j)}\|^2,
\label{eq:optimization_caricature}
\end{equation}
where the definitions of $\mathcal{N}_i$ and $c_{ij}$ are same as Eq.~\eqref{eq:vertex_representation}, $\mathcal{V}$ is the vertex set of the mesh. Since Eq.~\eqref{eq:optimization_caricature} is a non-linear least squares problem, we first compute the Jacobian matrix $\partial\mathbf{T}_{i}(\mathbf{w})/\partial\mathbf{w}$, and then use the Levenberg-Marquardt algorithm~\cite{more1978levenberg} to solve it.

For each 3D caricature from the constructed dataset in Sec.~\ref{sec:data}, we receive its corresponding optimal weight by solving Eq.~\eqref{eq:optimization_caricature}. Then we calculate the deformation representation of its every vertex via Eq.~\eqref{eq:DRCombination}. In our experiments, we select the reference mesh and $n$ deformed meshes from FaceWareHouse dataset~\cite{cao2013facewarehouse} ($n=99$). In detail, we choose the mean face as the reference mesh, $24$ expressions and $75$ identities with the neutral expression, which have large differences to the mean face, as the deformed meshes.

As the deformation representation for each vertex $[\mathbf{r}_i, \mathbf{s}_i]$ contains $9$ variables, the deformation representation of a whole 3D caricature mesh with $n_v$ vertices can be represented as a $9n_v$ vector $\{[\mathbf{r}_i,\mathbf{s}_i], i = 1,\ldots,n_{v}\}$. Based on the constructed 3D caricature dataset, we build a deformation space for 3D caricatures, where each 3D caricature is represented as a deformation form. Therefore, we formulate automatic caricature reconstruction as a geometric deformation problem, where the deformation representation of a 3D caricature is learned from a data-driven approach. In detail, given a 2D caricature, we aim to train an encoder-decoder framework that ends with several fully connected layers to regress its corresponding $9n_v$ deformation representation vector directly. Owing to the representation's robust expression ability, the translation from 2D caricature domain to 3D deformation field is quite natural.

\subsection{Landmark Detection and 3D Reconstruction}
\label{sec:network}
Although the deformation representation of 3D caricatures is well constructed, the large number of variables (each representation is a $9n_v$ vector) makes it hard for a convolutional neural network to regress the vector directly. To reduce the prediction difficulty of the network regressing the 3D shape, we make dimensionality reduction based on deformation representation. A similar approach is adopted in ~\cite{huang2014sparse}, which constructs the sparse localized basis of a triangle based deformation representation. However, different from~\cite{huang2014sparse}, we aim to estimate the deformation from the reference face to an arbitrary caricatured form, which need to capture the global shape deformation. Taking this into consideration, we adopt PCA model to assist network learning. Specifically, we propose an encoder-decoder framework to recover the 3D face shape and weak perspective parameters from the input 2D caricature image. We utilize the constructed PCA basis to initialize the last fully connected (FC) layer's weight. Based on the learnable statistical model, we propose a fully data-driven approach to this problem.

In detail, we propose a CNN-based approach to directly regress the intrinsic deformation representation and the weak perspective projection parameters with a single 2D caricature image. As shown in Fig.~\ref{fig:pipeline}, we utilize ResNet-34 backbone~\cite{he2016deep} to encode the input 2D caricature into a latent vector $\bm{\chi}\in \mathbb{R}^{216}$. The latent vector contains two parts, where $\bm{\chi}_{s}\in \mathbb{R}^{210}$ resolves the 3D shape and  $\bm{\chi}_{p}=(\hat{\mathbf{s}}, \hat{\mathbf{R}}, \hat{\mathbf{t}})\in \mathbb{R}^{6}$ represents the parameters of weak perspective projection, where the meanings of $\hat{\mathbf{s}}$, $\hat{\mathbf{R}}$, $\hat{\mathbf{t}}$ will be discussed later. We construct a decoder composed of $3$ fully connected layers to convert $\bm{\chi}_{s}$ to the estimated latent deformation representation $\{[\hat{\mathbf{r}}_{i},\hat{\mathbf{s}}_{i}], i = 1,\ldots,n_{v}\}$, where $n_{v}$ is the number of mesh vertices. The deformation gradients $\{(\log\hat{\mathbf{R}}_{i},\hat{\mathbf{S}}_{i}),i=1,\ldots,n_{v}\}$ and the deformation matrixes $\{\hat{\mathbf{T}}_{i},i=1,\ldots,n_{v}\}$ then can be recovered according to the derivation process in Sec.~\ref{sec:construction}. To help the model training, we use the first 500 principal components of a PCA basis extracted from the training dataset to initialize the weight of the last fully connected (FC) layer. \\

\noindent{\bf Loss for Caricature Shape.} As before, the estimated vertex coordinate $\{\hat{\mathbf{p}}_i'\}$ of target mesh can be obtained by solving
\begin{equation}
\mathop{\arg\min_{\{\hat{\mathbf{p}}_{i}^{\prime}\}}}\sum_{j\in{\mathcal{N}_i}} c_{ij}\|{(\hat{\mathbf{p}}_{i}^{\prime}-\hat{\mathbf{p}}_{j}^{\prime})-\hat{\mathbf{T}}_{i}(\mathbf{p}_{i}-\mathbf{p}_{j})}\|_{2}^2,
\end{equation}
which is equivalent to solve the following linear system:
\begin{equation}
2\sum_{j\in{\mathcal{N}_i}}c_{ij}(\hat{\mathbf{p}}_{i}^{\prime} - \hat{\mathbf{p}}_{j}^{\prime}) = \sum_{j\in{\mathcal{N}_i}}c_{ij}(\hat{\mathbf{T}}_{i} + \hat{\mathbf{T}}_{j})(\mathbf{p}_{i}-\mathbf{p}_{j}).
\end{equation}
As the deformation representation is translation independent, and thus we need to specify the position of mesh center or exactly one vertex. As the ground truth 3D caricature meshes are under the same specification, we construct a loss term to constrain the coordinate difference between the reconstructed mesh and the ground truth mesh as
\begin{equation}
\mathbf{E}_{ver}(\bm{\chi}_{s}) = \sum_{v_i\in\mathcal{V}} \|\hat{\mathbf{p}}_{i}^{\prime} - \mathbf{p}_{i}^{\prime}\|_{2}^2,
\end{equation}
where $\mathbf{p}_{i}^{\prime}$ presents the ground truth coordinate of the $i^{\rm th}$ vertex of the corresponding 3D mesh from the dataset, and $\mathcal{V}$ represents the vertex set.\\

\noindent{\bf Loss for Landmarks.} Reconstructing the 3D mesh from a 2D image is an inverse process of observing a 3D object by projecting it to 2D visual space. As before, we assume that the projection plane is the \emph{z}-plane and thus the scaled projection matrix can be written as $\mathbf{\Pi} =
s{	\left[ \begin{array}{ccc}
	1 & 0 & 0 \\
	0 & 1 & 0 \\
	\end{array} 
	\right ]},$ where $\mathbf{s}$ is the scale factor.
To better recover the landmark positions, we construct a landmark loss term to measure the difference between the projected landmarks and the ground truth landmarks:
\begin{equation}
\mathbf{E}_{lan}(\bm{\chi}_{s}, \bm{\chi}_{p}) = \sum_{v_i\in\mathcal{L'}}\|\hat{\mathbf{\Pi}}\hat{\mathbf{R}}\hat{\mathbf{p}}_{i}^{\prime}+\hat{\mathbf{t}}-\mathbf{q}_i'\|_{2}^2,
\end{equation}
where $\mathcal{L'}$ and $\mathcal{Q'} = \{\mathbf{q}_i',v_i\in\mathcal{L'}\}$ are the set of 3D landmarks and 2D landmarks separately, $\hat{\mathbf{\Pi}}$ is the estimated scaled projection matrix, $\hat{\mathbf{R}}$ is the estimated rotation matrix, and $\hat{\mathbf{t}}$ is the estimated translation vector. As our 3D caricature meshes have the same connectivities, the indices of 3D landmarks are the same for different caricature shapes.

\begin{figure}[!h]
	\begin{center}
		\includegraphics[width=1\linewidth]{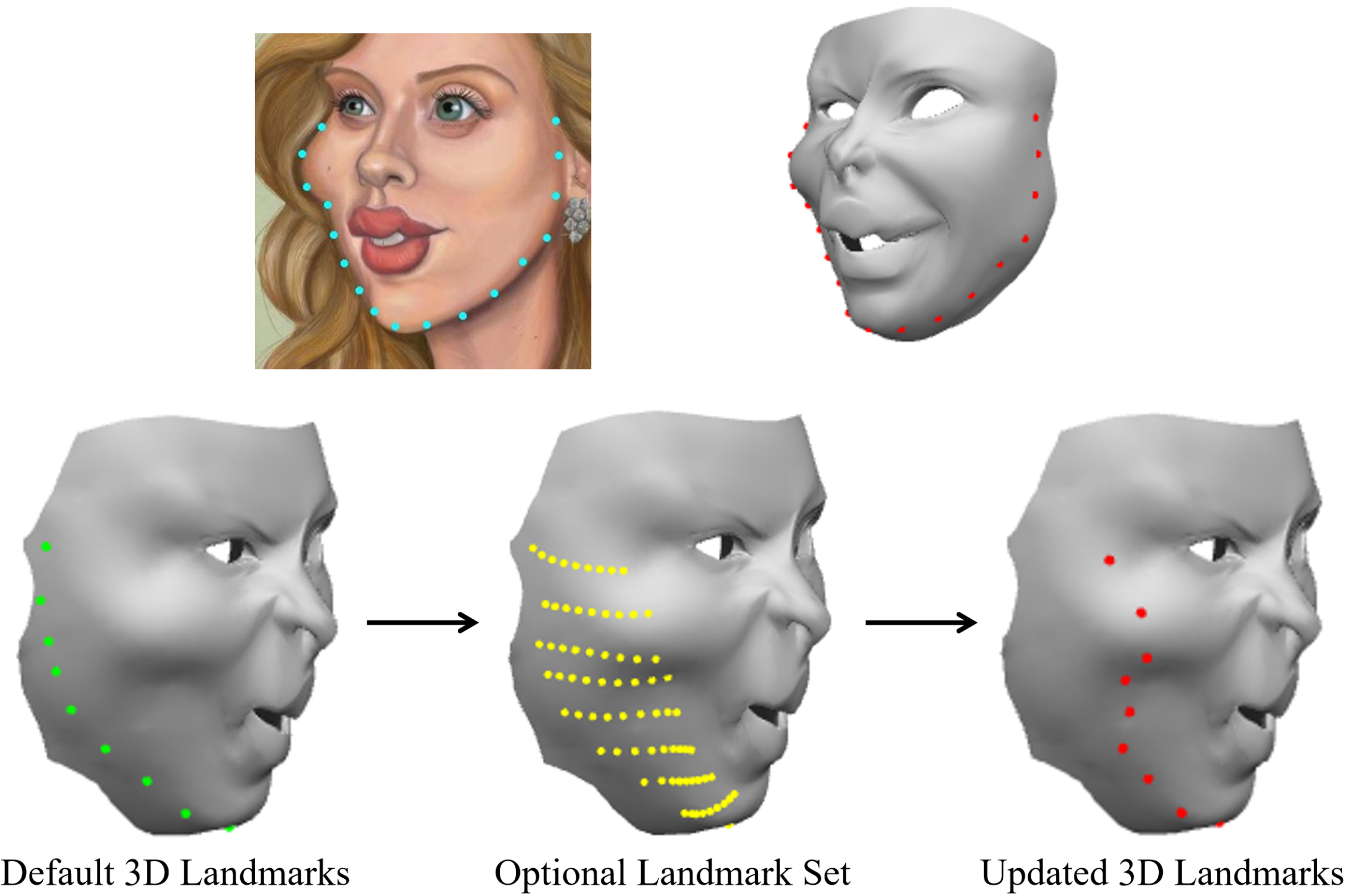}
	\end{center}
	\caption{For non-frontal face caricatures, we need to update the indices of silhouette landmarks on the 3D face shape to better match the corresponding 2D landmarks (shown in cyan in upper-left). The default 3D silhouette landmarks are shown in green in the lower-left. We construct an optional landmark set from each horizontal line (shown in yellow in lower-middle) that has a vertex lying on the silhouette and select among them a set of the updated silhouette landmarks according to the estimated rotation matrix $\hat{\mathbf{R}}$ in each training time. The vertices of the silhouette are updated in the end, as shown in red on the upper right and lower-right.}
	\label{fig:silhouette}
\end{figure}

Compared with normal face, the positions of caricature silhouette landmarks have large variance, and thus it is quite challenging to detect their positions accurately. Moreover, the 3D vertices corresponding with these silhouette landmarks are labeled on the mean neutral face with a frontal view, which causes the problem that the correspondences between 3D vertices and 2D landmarks are not correct for non-frontal faces as shown in Fig.~\ref{fig:silhouette}. To solve this problem, we update the indices of 3D silhouette landmarks during training according to the estimated rotation matrix and vertices' coordinates. In each training iteration, we select some vertices from each horizontal line that has a vertex lying on the silhouette and project them onto the image plane according to the estimated rotation matrix $\hat{\mathbf{R}}$. Then for each 2D silhouette landmark, we set the vertex whose projection is closet to it (see Fig.~\ref{fig:silhouette}) as its current corresponding 3D silhouette landmark.

The total loss function is given in the following form:
\begin{equation}
\begin{aligned}
\mathbf{E}= \lambda_1\mathbf{E}_{ver} + &\lambda_2\mathbf{E}_{lan},
\end{aligned}
\end{equation}
where $\lambda_1,\lambda_2$ are hyperparameters and their setting will be discussed in the experiment section.
\section{Experiments}

In this section, we give the implementation details, ablation studies, qualitative and quantitative evaluation of our proposed method, as well as comparisons with several related methods.\\

\noindent{\bf Implementation Details}
We train our model via the PyTorch~\cite{paszke2017automatic} framework. CNN takes the input of a color caricature image with size $224\times224\times3$. We use Adam solver~\cite{kingma:adam} with the mini-batch size of 32 and train the model with 2K iterations. The base learning rate is set to 0.0001. We set $\lambda_1 = 1, \lambda_2 = 0.00001$ during the first 1K iterations, and set $\lambda_1 = 1, \lambda_2 = 0.001$ during the last 1K iterations. The reason why the magnitudes of parameters are quite different is that the magnitude of vertices' coordinates has a big difference from that of 2D pixels.

All the tests, including our method and comparison methods, were conducted on a desktop PC with a hexa-core Intel CPU i7 at 3.40 GHz, 16GB of RAM, and NVIDIA TITAN Xp GPU. As for the running time for each caricature, our method takes about 10ms to obtain both 3D mesh and 68 2D landmarks. The number of vertices of our reconstructed mesh is 6144.

\subsection{Ablation Study}

We first conduct ablation studies to demonstrate the importance of each component. The ablation studies are designed for the augmented data, PCA initialization, and silhouette updating strategy.

As Tab.~\ref{table:ablation} shown, we evaluate the detection performance with several commonly used landmark error metrics, which are also used in~\cite{kowalski2017deep}. Specifically, the second row shows the error metrics of our method without using augmented data, which are generated by CariGANs~\cite{cao2018carigans} in Sec.~\ref{sec:data}. The third row shows the errors of our method without PCA initialization for the weight of the last FC layer mentioned in Sec.~\ref{sec:network}. And the fourth row shows the errors of our method without adopting the strategy to update the indices of silhouette landmarks displayed in Fig.~\ref{fig:silhouette}. As demonstrated in Tab.~\ref{table:ablation}, the mean error of estimated landmarks decreases from 5.85 to 5.64 with the help of augmented data, from 6.91 to 5.64 thanks to the PCA initialization, and from 5.99 to 5.64 owing to the silhouette updating strategy.

\begin{figure}[htb]
	\begin{center}
		\includegraphics[width=1.0\linewidth]{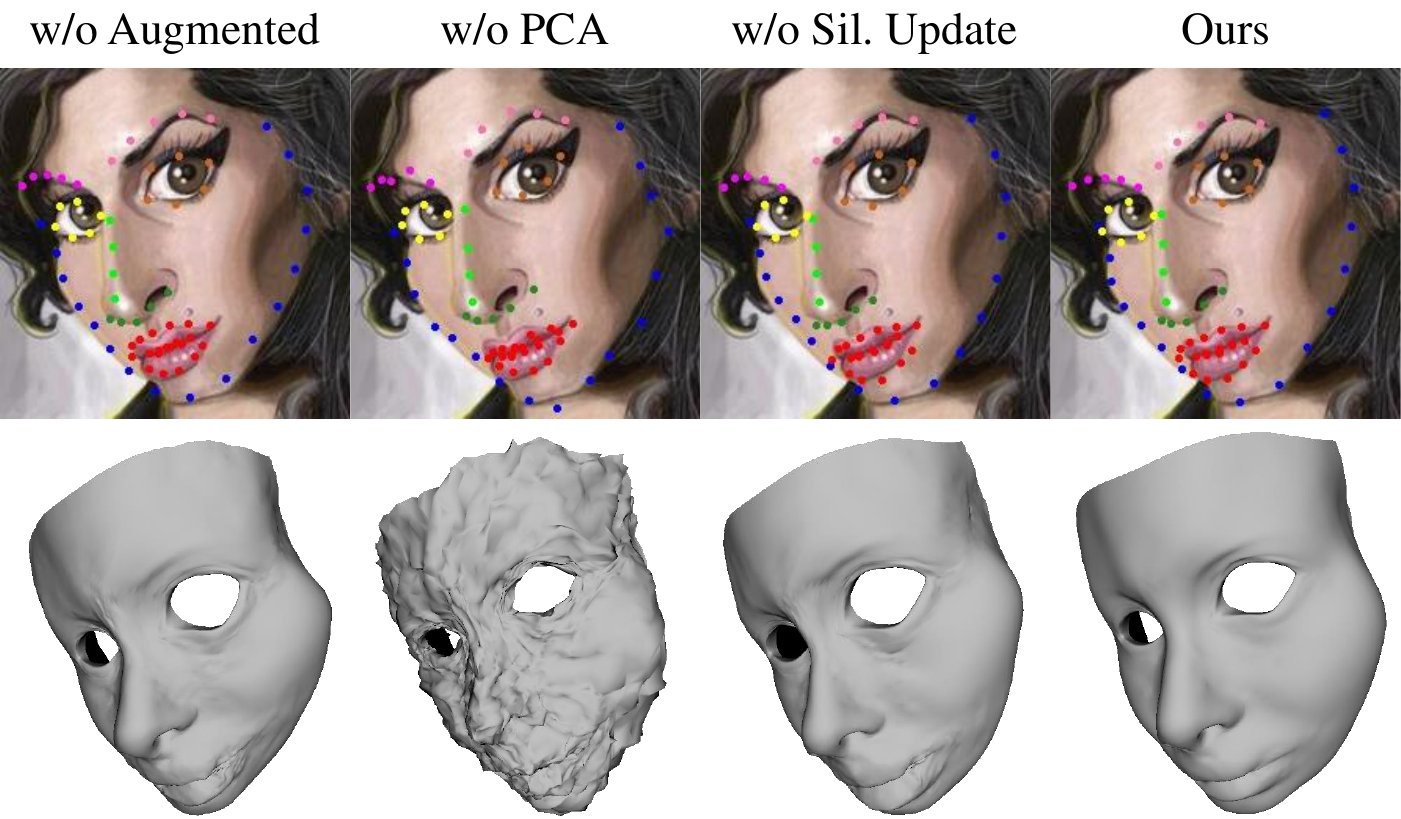}
	\end{center}
	\caption{Landmark detection and reconstruction results of the ablation studies. Left to right: results by method without using augmented data, results by method without PCA initialization, results by method without silhouette updating strategy, and results by our full method.}
	\label{fig:ablation}
\end{figure}

Fig.~\ref{fig:ablation} shows the detection and reconstruction results of ablation studies. The reconstructed mesh of our method without using augmented data shows that the learned model does not show good generalization ability, which leads to misalignment in the detection of silhouette points. The recovered mesh of our method without PCA initialization demonstrates that the PCA model makes the results to be smooth and more natural. We can observe that the predicted silhouette landmarks by the model without silhouette updating strategy deviate from the accurate position, which confirms the effectiveness of the silhouette updating strategy.

\begin{table}[htb]
	\centering
	\caption{Results of the ablation studies with metrics of landmark detection errors. Values of mean error with normalization are shown as the percentage of the normalization metric.}
	\newcommand{\tabincell}[2]{\begin{tabular}{@{}#1@{}}#2\end{tabular}}
	\begin{tabular}{|c|c|c|c|c|c|}
		\hline
		&\tabincell{c}{mean\\error}&\tabincell{c}{inter\\-pupil}&\tabincell{c}{inter\\-ocular}&diagonal \\
		\hline
		w/o Augmented&$5.85$&$9.29$&$6.34$&$2.38$ \\
		\hline
		w/o PCA&$6.91$&$11.01$&$7.52$&$2.82$ \\
		\hline
		w/o Sil. Update&$5.99$&$9.49$&$6.48$&$2.44$ \\
		\hline
		Ours&$\bm{5.64}$&$\bm{8.93}$&$\bm{6.10}$&$\bm{2.30}$ \\
		\hline
	\end{tabular}
	\label{table:ablation}
\end{table}

\subsection{Detection Comparison}

\begin{figure*}[htb]
	\begin{center}
		\includegraphics[width=1.0\linewidth]{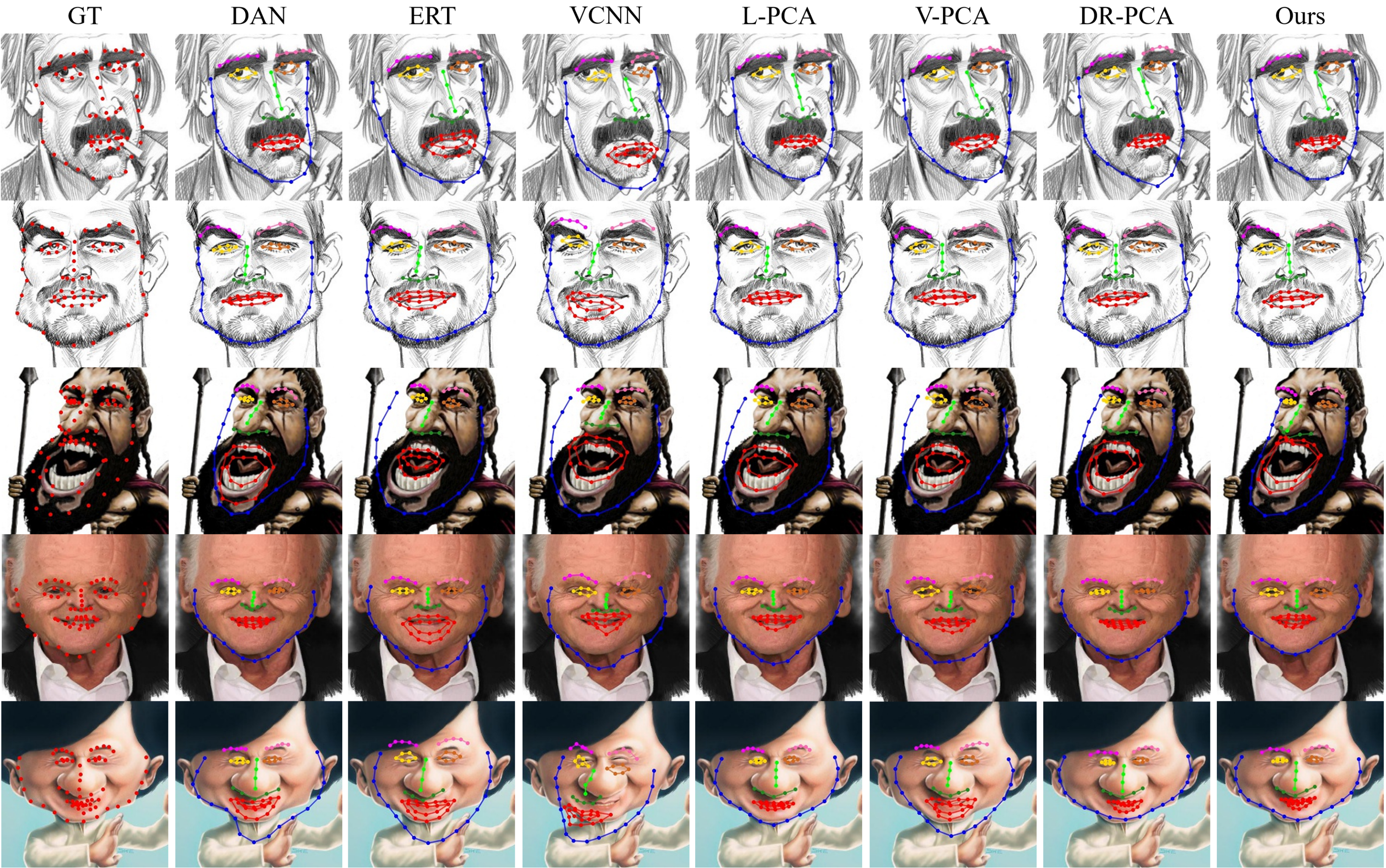}
	\end{center}
	\caption{We provide visual landmark detection results on the test dataset using DAN~\cite{kowalski2017deep}, ERT~\cite{kazemi2014one}, VCNN~\cite{wu2017facial} and some baselines, including Landmark PCA (L-PCA), Vertex PCA (V-PCA), and DR-PCA.}
	\label{fig:compare_land}
\end{figure*}

As far as we know, there is no existing method for landmark detection for general caricatures. We compare our method with some benchmark methods. The first type is the face alignment methods, which are designed for normal human faces, and we select three typical methods, including DAN~\cite{kowalski2017deep}, ERT~\cite{kazemi2014one}, and vanilla CNN (VCNN) designed by~\cite{wu2017facial}. As their released trained models are trained with normal facial images, we retrain their models based on the author's training code. For a fair comparison, their methods are trained and tested with the same training and testing dataset as our method. We randomly split our dataset into 80\% for training and 20\% for testing.

\noindent\textbf{DAN:} Deep Alignment Network (DAN)~\cite{kowalski2017deep} is a robust face alignment method based on deep neural network architecture. Its algorithm pipeline includes multiple stages, where each stage improves the locations of the facial landmarks estimated by the previous stage.

\noindent\textbf{ERT:} In~\cite{kazemi2014one}, an ensemble of regression trees (ERT) has been used to directly estimate the facial landmark positions from a sparse subset of pixel intensities. This method achieves super-realtime performance with high-quality predictions. It has been integrated into the Dlib library~\cite{king2009dlib}.

\noindent\textbf{VCNN:} Vanilla CNN is proposed in~\cite{wu2017facial}, which introduces hierarchical and discriminative processing to the existing CNN design for facial landmark regression.

Except for the above three methods, we also implement some baseline methods.

\noindent\textbf{L-PCA:} Inspired by~\cite{cao2018carigans}, we extract the PCA basis of 2D caricature landmarks from the labeled landmark dataset. In this way, the landmarks of caricature image can be represented by the coefficient of PCA basis. We use the same ResNet framework in our method to directly regress the coefficient.

\noindent\textbf{V-PCA:} We extract the PCA basis of 3D caricature shape set represented by the Euclidean coordinates. The network structure is the same as our algorithm pipeline in Fig.~\ref{fig:pipeline}, and regresses the PCA coefficient and orientation.

\noindent\textbf{DR-PCA:} We extract the first 210 principal components of a PCA basis from the 3D caricature shape set represented by the deformation representation. The pipeline is the same as our method by changing the decoder (the last 3 FC layers) to the matrix multiplication with the extracted PCA basis. Specifically, the deformation gradients $\{(\log\hat{\mathbf{R}}_{i},\hat{\mathbf{S}}_{i}),i=1,\ldots,n_{v}\}$ can be directly computed via matrix multiplication between the extracted PCA basis and the latent vector $\bm{\chi}_{s}\in \mathbb{R}^{210}$.

\begin{table}[htb]
	\centering
	\caption{Statistics of landmark detection errors and computation time (ms/image) on the test set. Values of mean error with normalization are shown as the percentage of the normalization metric.}
	\newcommand{\tabincell}[2]{\begin{tabular}{@{}#1@{}}#2\end{tabular}}
	\begin{tabular}{|c|c|c|c|c|c|c|}
		\hline
		&\tabincell{c}{mean\\error}&\tabincell{c}{inter\\-pupil}&\tabincell{c}{inter\\-ocular}&diagonal&\tabincell{c}{time\\(ms)} \\
		\hline
		DAN&$5.78$&$9.93$&$6.80$&$2.59$&$25.9$ \\
		\hline
		ERT&$8.24$&$14.52$&$9.95$&$3.71$&$2.7$ \\
		\hline
		VCNN&$14.04$&$24.33$&$16.67$&$6.39$&$\bm{1.6}$ \\
		\hline
		L-PCA&$5.87$&$10.08$&$6.91$&$2.64$&$4.8$ \\
		\hline
		V-PCA&$6.20$&$10.68$&$7.32$&$2.79$&$6.4$ \\
		\hline
		DR-PCA&$5.75$&$9.89$&$6.77$&$2.58$&$9.3$\\
		\hline
		Ours&$\bm{4.98}$&$\bm{8.51}$&$\bm{5.82}$&$\bm{2.23}$&$9.8$\\
		\hline
	\end{tabular}
	\label{table:fitting}
\end{table}

\begin{figure}
	\begin{center}
		\includegraphics[width=1.0\linewidth]{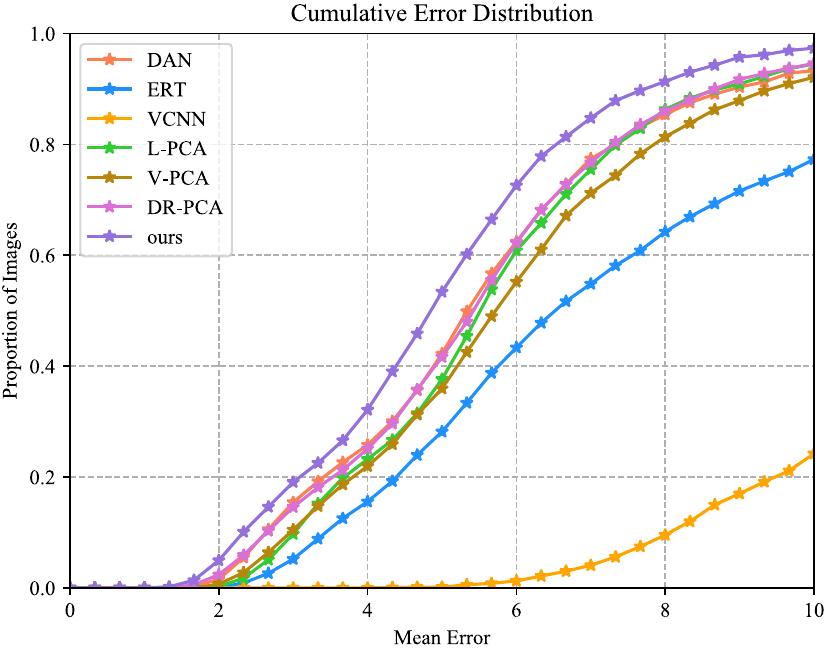}
	\end{center}
	\caption{Comparisons of cumulative errors distribution (CED) curves on the test set.}
	\label{fig:CED}
\end{figure}

We compare our method with these benchmark methods. Fig.~\ref{fig:compare_land} shows some visual results of landmark detection. It can be observed that the detected landmarks of ERT~\cite{kazemi2014one} and VCNN~\cite{wu2017facial} can not match the face shape. The method of DAN~\cite{kowalski2017deep} performs quite well for the facial feature parts, including eyes, nose, and mouth. However, its silhouette landmarks may deviate from the accurate positions. V-PCA and L-PCA are also not good for the landmarks on the silhouette. Though DR-PCA representation shows nice performance, it still can not match the facial feature parts precisely. In contrast, the detection results by our method are quite close to the ground truth landmarks, even for the silhouette landmarks. We also quantitatively compare our method with these methods on several frequently used landmark error metrics and average computation time. We show the statistics in Tab.~\ref{table:fitting} and the cumulative errors distribution (CED) curves of these methods on the mean error in Fig.~\ref{fig:CED}. We can see that the mean error, mean error normalized separately by inter-pupil distance, inter-ocular distance, and bounding box diagonal of our methods are all smaller than those of other methods.

The reason why our method performs better includes the following aspects. First, rather than directly regressing the 2D landmarks, we regress the 3D shape and orientation. In this way, a challenging problem is decomposed into two easier problems. Second, to better represent the 3D caricature shape, we learn a nonlinear parametric model, which is more suitable to represent the 3D caricature shape than 3D morphable model~\cite{blanz1999morphable} and FaceWareHouse~\cite{cao2013facewarehouse}.

\subsection{3D Reconstruction Comparison}

\begin{figure}
	\begin{center}
		\includegraphics[width=1.0\linewidth]{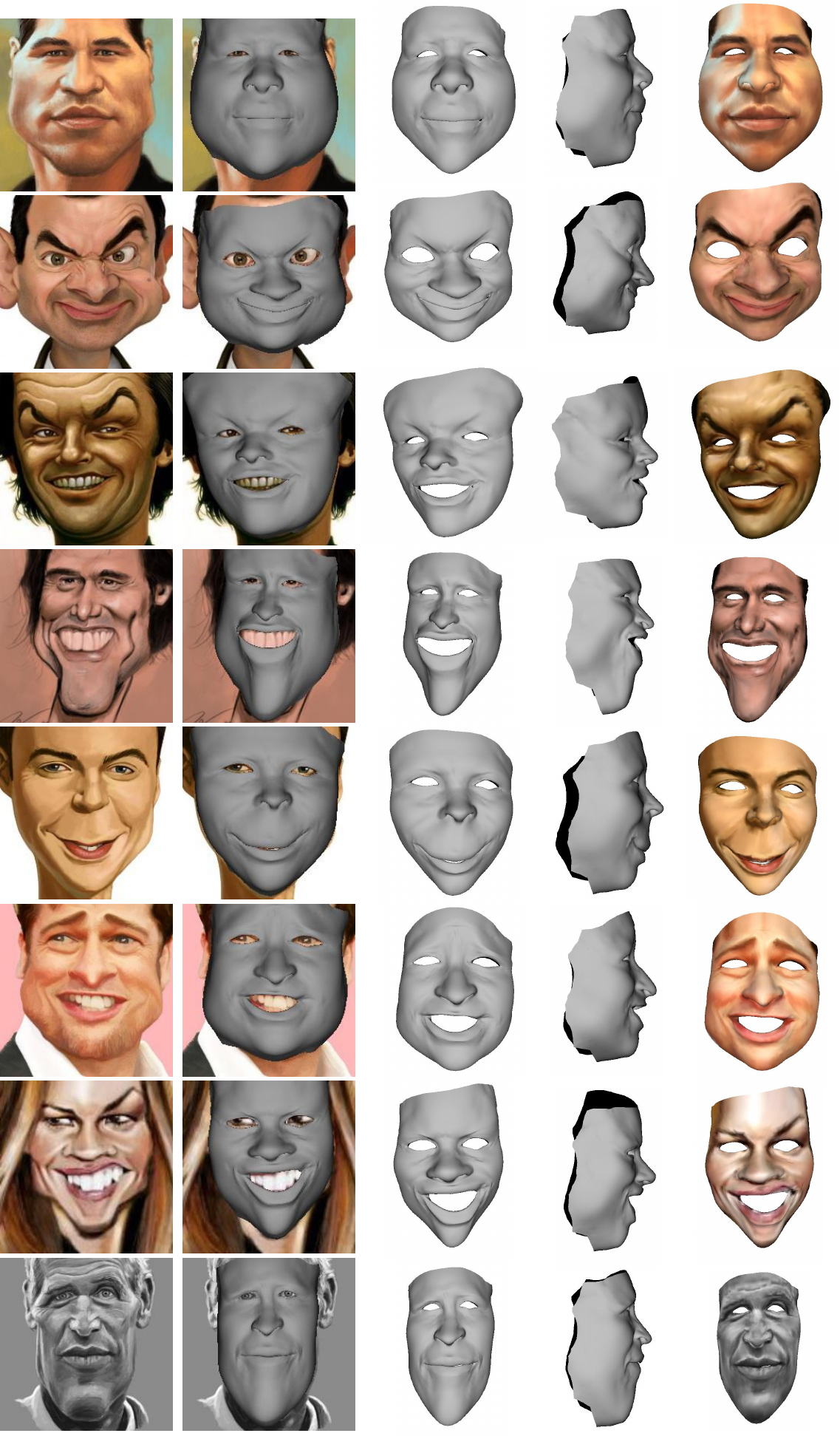}
	\end{center}
	\caption{Left to right: input caricature, predicted mesh overlaying on the image, predicted mesh in two different views, predicted mesh with texture.}
	\label{fig:texture}
\end{figure}

Reconstructing 3D caricature shape from caricature image is also a challenging problem. In Fig.~\ref{fig:texture}, we show eight reconstruction examples from the test set. The reconstructed mesh is overplayed on the image, and we can observe that the shape is recovered quite well. The recovered mesh from two different views and with texture are also shown to demonstrate the effectiveness of our method.

\begin{figure}
	\begin{center}
		\includegraphics[width=1.0\linewidth]{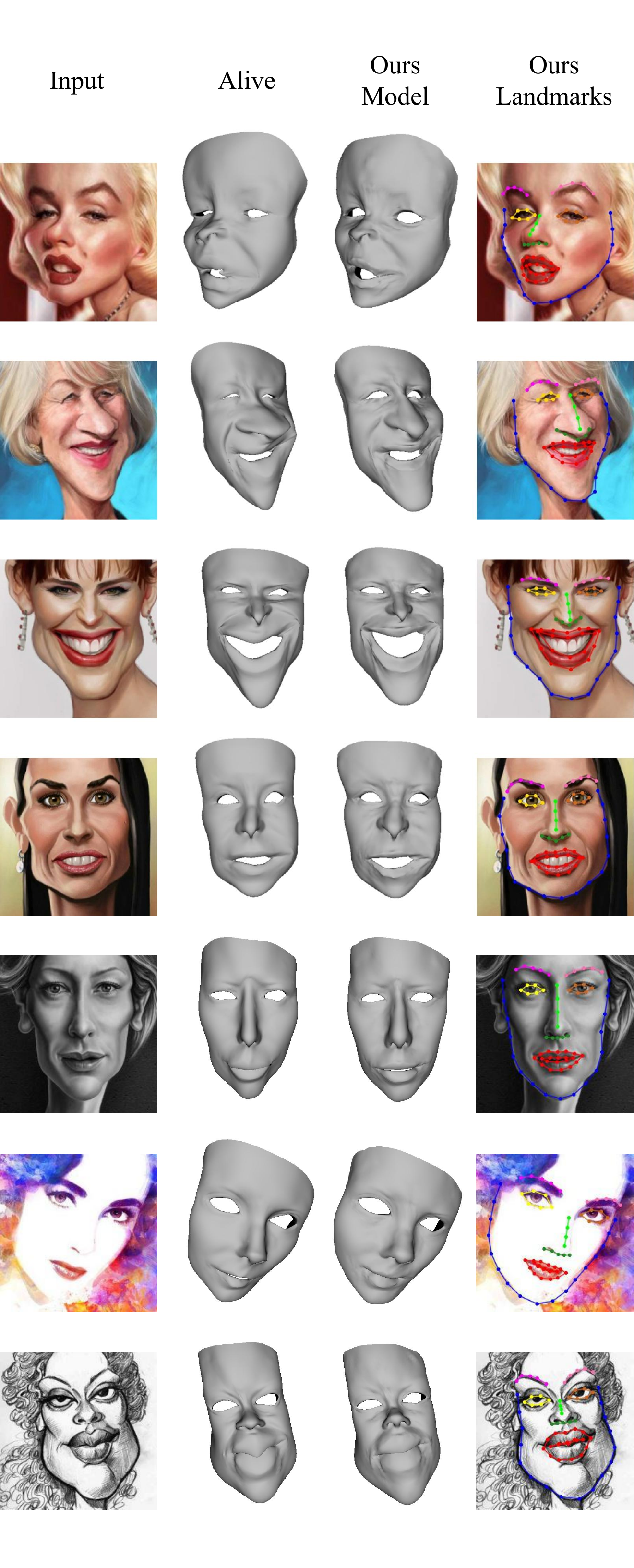}
	\end{center}
	\caption{Reconstruction results by our method and~\cite{wu2018alive} which needs labeled landmarks. From the first column to the last column are input images, reconstruction results by~\cite{wu2018alive}, reconstruction results by our method, and the projected 2D landmarks by our method, respectively.}
	\label{fig:compare_alive}
\end{figure}

We also compare our method with an existing state-of-the-art method~\cite{wu2018alive}, which is the only universal method of 3D caricature reconstruction. Then as shown in their paper, classical parametric models like 3DMM~\cite{paysan20093d,zhu2015high} and FaceWareHouse~\cite{cao2013facewarehouse} cannot reconstruct exaggerated meshes well due to their limited extrapolation ability. As Fig.~\ref{fig:compare_alive} shows, compared to the results of~\cite{wu2018alive}, the reconstructed 3D meshes by our method are quite natural and vivid. There are two advantages of our method. One is the computation time. It takes around 10ms (real-time) to produce the result with our method, while 12.5s for their method. Another difference is that our reconstruction method does not need to label the landmarks manually, while~\cite{wu2018alive} needs labeled landmarks as input.

Moreover, we also compare our method with the reconstruction methods by 3DMM~\cite{zhu2015high}, FaceWareHouse~\cite{cao2013facewarehouse,cao2014displaced} and Alive~\cite{wu2018alive}. We compare the mean square error between the projected landmarks and ground-truth landmarks, which also has been used in~\cite{wu2018alive}. For these optimization based methods, the 3D caricature mesh is reconstructed by minimizing the residuals between the projected landmarks and the ground-truth landmarks. It needs to be pointed out that the compared methods all need labeled landmarks input, while our method is automatic. As shown in Tab.~\ref{table:error}, our method even outperforms both 3DMM and FaceWareHouse fitting over test data, although our method does not have the ground truth 2D caricature landmarks as input.

\begin{table}[htb]
	\centering
	\caption{The mean square error between projected landmarks and ground-truth landmarks over test data. The first row shows the methods, and the second row shows their corresponding mean square errors of landmarks. Note that the compared methods all need labeled landmarks input, while our method automatically detects the landmarks and reconstructs the 3D mesh from the input caricature.}
	\newcommand{\tabincell}[2]{\begin{tabular}{@{}#1@{}}#2\end{tabular}}
	\begin{tabular}{|c|c|c|c|c|c|c|}
		\hline
		3DMM&FaceWareHouse&Alive&Ours \\
		\hline
		$6.32$&$7.61$&$0.02$&$4.98$ \\
		\hline
	\end{tabular}
	\label{table:error}
\end{table}

\begin{figure*}
	\begin{center}
		\includegraphics[width=0.85\linewidth]{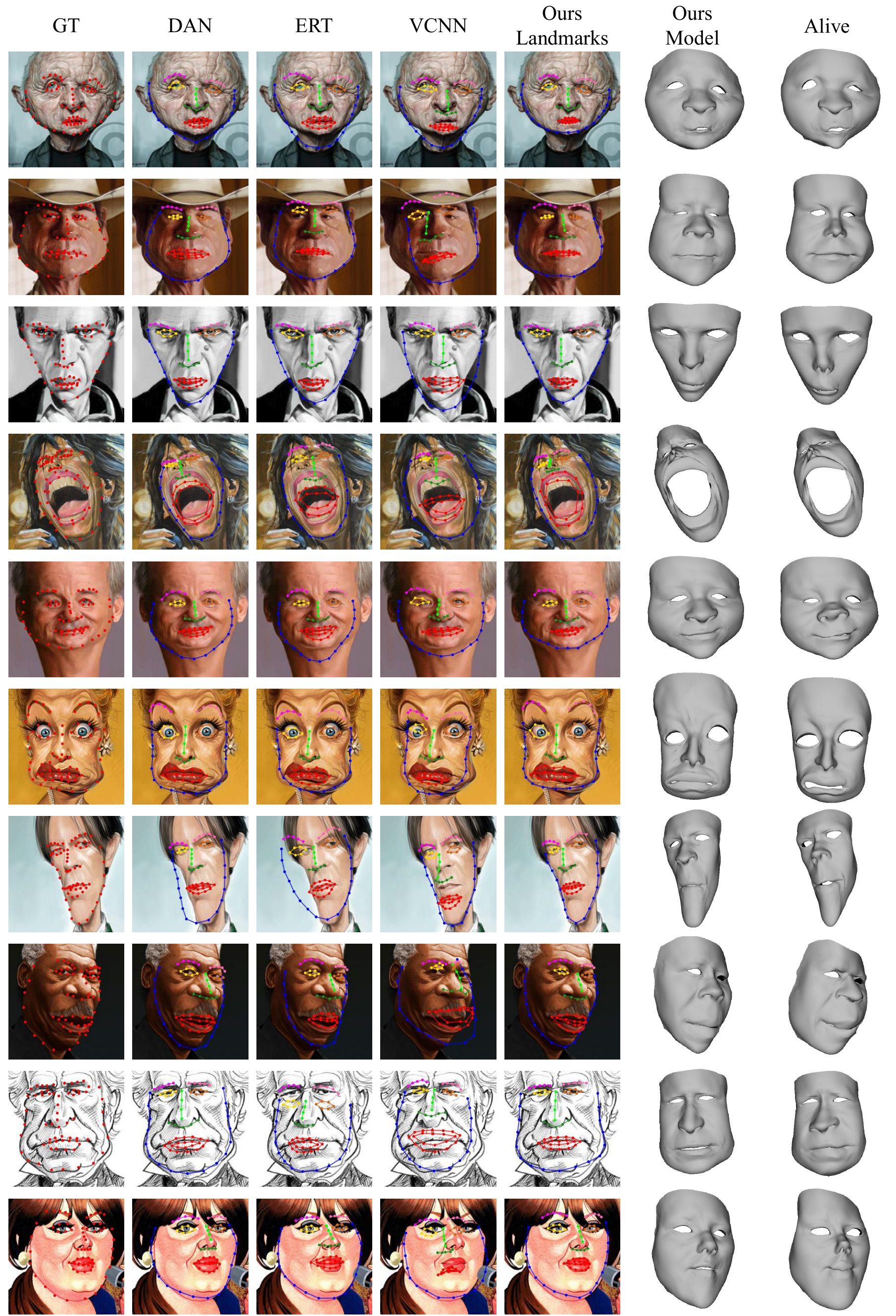}
	\end{center}
	\caption{Landmark detection comparisons with benchmark methods DAN~\cite{kowalski2017deep}, ERT~\cite{kazemi2014one}, VCNN~\cite{wu2017facial} and reconstruction comparisons with state-of-the-art method ~\cite{wu2018alive} which needs labeled landmarks. It can be seen that our method can detect landmarks and reconstruct 3D face shapes quite well.}
	\label{fig:compare1_all}
\end{figure*}

From the above quantitative and qualitative experiments, we can see that our proposed method performs quite well on landmark detection and reconstruction for caricatures. In Fig.~\ref{fig:compare1_all}, more experimental results and comparisons with the benchmark landmark detection methods~\cite{kowalski2017deep,kazemi2014one,wu2017facial} and the existing state-of-the-art caricature reconstruction method~\cite{wu2018alive} on $10$ test caricatures are given. These results further validate the superior effect of our proposed method on the tasks of landmark detection and 3D reconstruction on caricature.

\section{Conclusion}
We have presented an effective and efficient algorithm for automatic landmark detection and 3D reconstruction for 2D caricature images. This challenging problem is well solved by separately regressing the 3D face shape and face pose, and then 2D landmarks and 3D shape can both be obtained. To represent the non-regular 3D caricature face, we construct a 3D caricature shape dataset to learn the latent representation. Extensive experimental results show that the detected 2D landmarks and reconstructed 3D face shape fit the caricature quite well, which outperforms the existing state-of-the-art methods in both computation speed and accuracy.

\noindent \textbf{Acknowledgement}
This work was supported by the National Natural Science Foundation of China (No. 61672481) and Youth Innovation Promotion Association CAS (No. 2018495).

\bibliographystyle{IEEEtran}
\bibliography{egbib}

\end{document}